\documentclass[letterpaper]{article} 
\usepackage{aaai2026}  
\usepackage{times}  
\usepackage{helvet}  
\usepackage{courier}  
\usepackage[hyphens]{url}  
\usepackage{graphicx} 
\usepackage{natbib}  
\usepackage{caption} 
\usepackage{algorithm}
\usepackage{algorithmic}

\usepackage{newfloat}
\usepackage{listings}
\DeclareCaptionStyle{ruled}{labelfont=normalfont,labelsep=colon,strut=off} 
\floatstyle{ruled}
\newfloat{listing}{tb}{lst}{}
\floatname{listing}{Listing}
\usepackage{amsmath}
\usepackage{booktabs}
\usepackage{multirow}
\title{N2N-GQA: Noise-to-Narrative for Graph-Based Table-Text Question Answering Using LLMs\thanks{
		This arXiv version reflects a corrected author list following an authorship oversight during submission.
		}}

\author{
	Mohamed Sharafath\equalcontrib\textsuperscript{\rm 1},
	Aravindh Annamalai\equalcontrib\textsuperscript{\rm 1}\thanks{Work done during internship at Comcast India Engineering Center (CIEC).},
	Ganesh Murugan\textsuperscript{\rm 2},
	Aravindakumar Venugopalan\textsuperscript{\rm 3}
}
\affiliations{
	\textsuperscript{\rm 1}Comcast India Engineering Center, Chennai, India\\
	\textsuperscript{\rm 2}Comcast, Philadelphia, United States\\
 mmoham795@apac.comcast.com, aravindhannamalai\_g@comcast.com,\\ ganesh\_murugan@comcast.com, aravindakumar\_venugopalan@cable.comcast.com
}

\begin{document}
	
	\maketitle
	
	\begin{abstract}
		Multi-hop question answering over hybrid table-text data requires retrieving and reasoning across multiple evidence pieces from large corpora, but standard Retrieval-Augmented Generation (RAG) pipelines process documents as flat ranked lists, causing retrieval noise to obscure reasoning chains. We introduce N2N-GQA. To our knowledge, it is the first zero-shot framework for open-domain hybrid table-text QA that constructs dynamic evidence graphs from noisy retrieval outputs. Our key insight is that multi-hop reasoning requires understanding relationships between evidence pieces: by modeling documents as graph nodes with semantic relationships as edges, we identify bridge documents connecting reasoning steps, a capability absent in list-based retrieval. On OTT-QA, graph-based evidence curation provides a 19.9-point EM improvement over strong baselines, demonstrating that organizing retrieval results as structured graphs is critical for multi-hop reasoning. N2N-GQA achieves 48.80 EM, matching fine-tuned retrieval models (CORE: 49.0 EM) and approaching heavily optimized systems (COS: 56.9 EM) without any task-specific training. This establishes graph-structured evidence organization as essential for scalable, zero-shot multi-hop QA systems and demonstrates that simple, interpretable graph construction can rival sophisticated fine-tuned approaches.
	\end{abstract}
	
	\section{Introduction} Real-world question answering (QA) often requires reasoning over multiple pieces of information spread across diverse data formats, such as structured tables and unstructured text. Such multi-hop questions are challenging because they require a system to find an initial piece of evidence and use it to find subsequent, related facts to synthesize a final answer. The initial retrieval step is critical; errors made here can propagate through the reasoning chain, causing the entire query to fail. While standard Retrieval-Augmented Generation (RAG) pipelines have advanced QA, they often treat retrieved documents as a simple, unstructured list, failing to capitalize on the relationships between evidence pieces. This ``retrieval noise'' can easily overwhelm a downstream reader model, especially in complex reasoning tasks.
	
	The core insight of our work is that multi-hop reasoning requires understanding relationships between evidence pieces, not just their individual relevance. Consider a question like ``What is the capital of the country where the 2019 Atlantic Hockey Player of the Year was born?'' This requires: (1) identifying the player, (2) finding their birthplace country, and (3) determining that country's capital. Standard retrieval returns documents scored by individual relevance to the question, but cannot identify which documents form a connected reasoning chain.
	
	Unlike closed-domain settings where gold evidence is provided, open-domain hybrid QA requires first retrieving relevant tables and passages from large corpora, a challenge that has previously required extensive fine-tuning COS \citep{ma2023chain}. Existing zero-shot methods operate in closed-domain settings with gold evidence ODYSSEY \citep{agarwal2025hybrid}, leaving a critical gap for deployable, training-free systems.
	
	Graph structures naturally represent these connections. By modeling retrieved documents as nodes and their semantic relationships as edges, we can identify documents that are both individually relevant and structurally important serving as bridges between reasoning steps. This ``connectivity-aware'' evidence selection is fundamentally different from ranked-list retrieval, where each document is evaluated in isolation.
	
	To address this, we introduce N2N-GQA (Noise-to-Narrative for Graph-based Question Answering), a zero-shot framework that transforms noisy retrieved documents into a structured evidence graph. This ``noise-to-narrative'' process constructs and prunes a query-specific graph, filtering noise to identify the core reasoning path and the "bridge" documents that connect it. As a refinement, we also introduce GraphRank, an algorithm combining semantic relevance with graph centrality. Our contributions are: \begin{itemize} \item A systematic demonstration that \textbf{graph-based evidence curation} organizing retrieved documents as connected knowledge structures provides dramatic improvements over list-based retrieval, establishing graph organization as essential for multi-hop reasoning. \item A novel, zero-shot framework (N2N-GQA) that transforms noisy search results into a coherent context narrative via dynamic graph construction and curation. \item GraphRank, a refinement technique combining semantic relevance with graph centrality, providing consistent modest improvements as a complementary enhancement. \item Strong empirical evidence that zero-shot graph-based RAG can approach fine-tuned model performance (48.80 EM on OTT-QA), offering a more scalable and adaptable paradigm for multi-hop QA systems. \end{itemize}
	
	In our framework, the evidence graph acts as an explicit symbolic structure that organizes retrieved information into a form that the LLM can reason over more reliably than with an unstructured document list.

	\section{Related Work}
	\subsection{Multi-hop Question Answering} Multi-hop question answering (QA) requires integrating information from multiple sources. Benchmarks have evolved from purely textual reasoning tasks, such as HotpotQA \citep{yang2018hotpotqa}, to hybrid table-text formats like HybridQA \citep{chen2020hybridqa}. A key challenge lies in the open-domain setting, exemplified by OTT-QA \citep{chen2020open}, which requires retrieval from a large corpus, unlike closed-domain tasks such as HybridQA \citep{chen2020hybridqa} where gold evidence is provided.
	
	Early approaches focused on fine-tuning pipeline systems of retrievers and readers like HybridQA, MuGER, S3HQA \citep{chen2020hybridqa, wang2022muger, lei2023s}. However, top-performing systems in the open-domain setting such as CARP \citep{zhong2022reasoning}, CORE \citep{ma2022open}, and COS \citep{ma2023chain} require extensive, task-specific pre-training and fine-tuning to build and reason over evidence chains or graphs.
	
	Recent work has also explored using graph structures with LLMs. For instance, ODYSSEY construct ``Hybrid Graphs'' from predefined schemas, demonstrating that graph representations improve answer quality in closed-domain settings. Our work differs fundamentally as we \textbf{dynamically construct graphs from noisy retrieval outputs}. The graph structure emerges from the retrieved documents rather than being predefined. This makes our approach applicable to open-domain settings where comprehensive knowledge graphs are unavailable and no fine-tuning is performed. Our empirical results validate that this dynamic graph-structured evidence organization is critical for multi-hop reasoning.
	
	\subsection{Zero-Shot vs. Fine-Tuned Open-Domain QA}
	
	The open-domain setting presents fundamental challenges absent in closed-domain work. While methods like ODYSSEY demonstrate strong zero-shot reasoning given gold evidence, they do not address the retrieval problem, the primary bottleneck in open-domain deployment.
	
	Such graph-based representations also function as symbolic structures that help constrain and organize the reasoning process of large language models.
	
	For open-domain hybrid QA, all prior competitive systems require extensive fine-tuning:
	\begin{itemize}
		\item CORE trains DPR-based retrievers with entity linking.
		\item COS pre-trains on Wikipedia-scale data.
	\end{itemize}
	
	To our knowledge, N2N-GQA is the first zero-shot framework for open-domain hybrid table-text QA, providing a practical alternative to these training-intensive approaches.
	
	\begin{figure*}[t]
		\centering
		\includegraphics[width=0.9\textwidth]{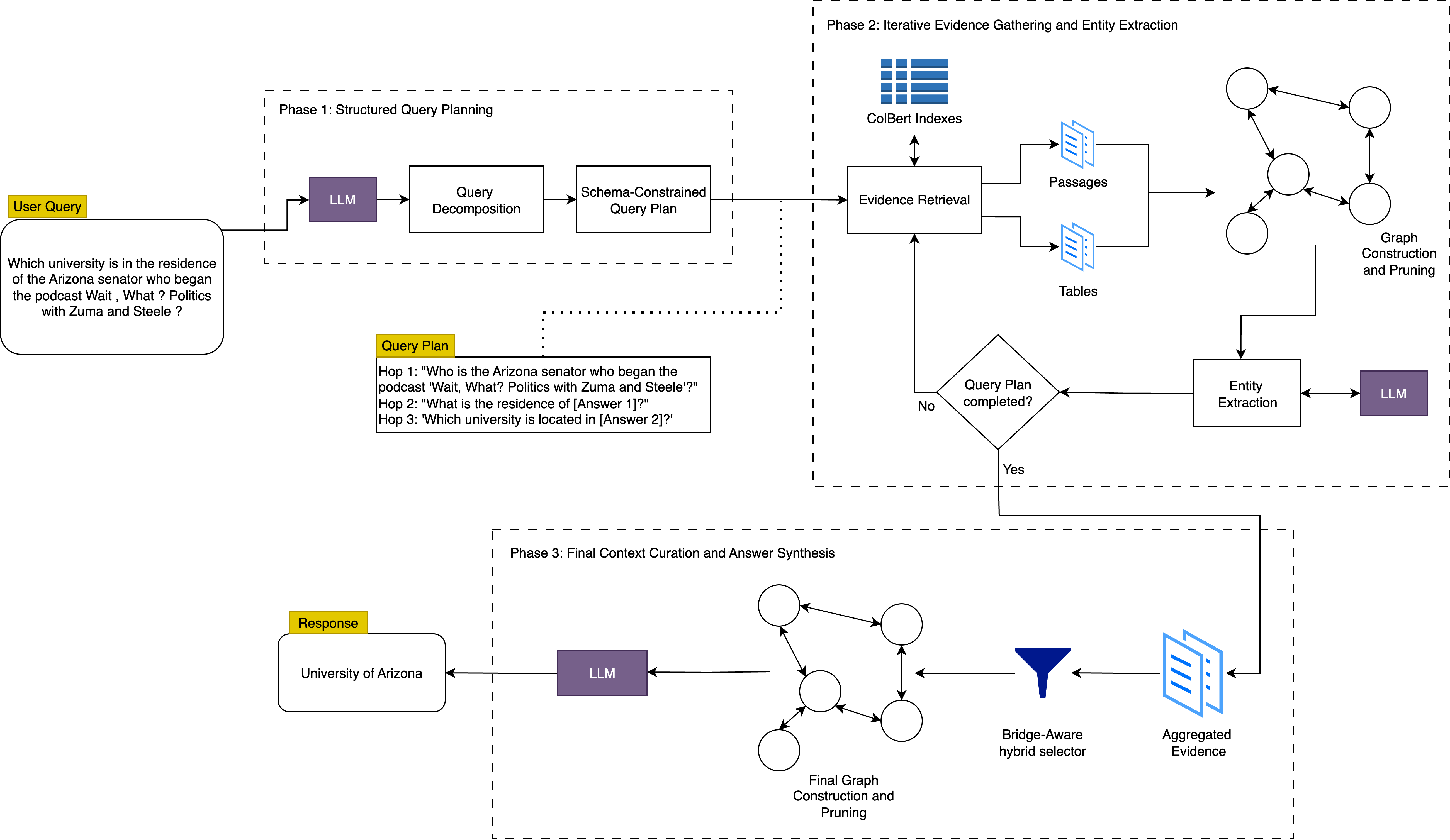}
		\caption{\textbf{Overview of N2N-GQA.} The pipeline begins with structured query planning, followed by iterative evidence gathering where documents are modeled as a dynamic graph. The ``noise-to-narrative'' process uses centrality-based pruning to identify bridge documents, resulting in a curated evidence graph for final answer synthesis.}
		\label{fig:architecture}
	\end{figure*}
	
	\section{Methodology: Graph-Based Retrieval Augmentation}
	The N2N-GQA pipeline is a zero-shot, fine-tuning-free system designed for multi-hop question answering over hybrid data. It processes a user's query through a multi-stage architecture that includes query planning, iterative evidence gathering and entity extraction, final context curation, and answer synthesis.
	
	\subsection{Problem Definition}
	The fundamental task of multi-hop question answering over hybrid data is to find a correct answer, $A$, for a complex natural language question, $Q$, by synthesizing evidence from a large, heterogeneous corpus, $C$. This corpus consists of a set of unstructured text passages, $\mathcal{P}$, and a set of structured, serialized table rows  , $\mathcal{T}$, such that $C = \mathcal{P} \cup \mathcal{T}$. Formally, a multi-hop question, $Q$, implies a reasoning process that cannot be resolved by a single document retrieval. Instead, it requires the identification of a sequence of intermediate entities $(e_1, e_2, \dots, e_{n-1})$ to arrive at the final answer, $A=e_n$. Answering $Q$ is therefore equivalent to finding an optimal reasoning chain, $\mathcal{R}$, which is an ordered sequence of evidence documents $(d_1, d_2, \dots, d_n)$, where each $d_i \in C$. Our work models the set of candidate documents retrieved for $Q$ as a knowledge graph, $G = (V, E)$, where the set of vertices $V = \{v_1, v_2, \dots, v_k\}$ represents the candidate documents, and the set of weighted edges $E$ represents the relationships between them. The core problem is to identify an optimal subgraph, $G^* \subseteq G$, that represents the most coherent and relevant evidence for answering $Q$. 
	
	\subsubsection{Evidence Graphs as Dynamic Knowledge Structures}
	Our evidence graph differs from traditional knowledge graphs in important ways. Rather than representing general world knowledge with entities and predefined relations, our graph is \textbf{query-specific and dynamically constructed} from retrieval results. Each node $v \in V$ represents a retrieved document (passage or serialized table row), and each edge $e \in E$ represents semantic overlap between documents, weighted by shared term importance.
	
	This construction serves two purposes:
	\begin{enumerate}
		\item \textbf{Relationship Discovery}: Documents that share important terms (high TF-IDF overlap) likely discuss related entities or concepts, forming potential links in a reasoning chain. Edge weights quantify these relationships.
		\item \textbf{Structural Importance}: Graph centrality metrics identify documents that connect multiple pieces of evidence ``hub'' documents that are referenced by or reference many other documents. These are often critical bridge documents for multi-hop reasoning.
	\end{enumerate}
	
	Crucially, our graph construction is \textbf{unsupervised and deterministic} requiring no training data. It transforms neural retrieval outputs (ranked documents) into structured knowledge representations that enable graph-theoretic reasoning operations. This approach scales to any domain where retrieval is possible, without requiring domain-specific knowledge graphs.
	
	\subsubsection{GraphRank Scoring Function}
	We define a ranking function, $Score_{GR}(v)$, for each node $v \in V$. To ensure stable and predictable ranking, we first normalize both the semantic and structural scores to a [0, 1] range using min-max scaling across all nodes in the current graph. The final score is then calculated using a multiplicative approach:
	\begin{equation}
		Score_{GR}(v) = S_{sem\_norm}(v) \times (1 + (1 - \alpha) \times S_{struct\_norm}(v))
		\label{eq:graphrank}
	\end{equation}
	where:
	\begin{itemize}
		\item $S_{sem\_norm}(v)$ is the normalized semantic similarity score for document $v$.
		\item $S_{struct\_norm}(v)$ is the normalized weighted centrality of node $v$ within the graph $G$.
		\item $\alpha \in [0, 1]$ is a hyperparameter that balances the influence of the structural score, which acts as a confidence booster for the primary semantic score.
	\end{itemize}
	
	\textbf{Design Rationale.} The multiplicative structure in Equation~\ref{eq:graphrank} enforces \textbf{semantic dominance}: $S_{sem\_norm}$ acts as a gating function, ensuring that structurally central but semantically irrelevant nodes are suppressed, a distinct advantage over additive formulations where high centrality can introduce noise. Consequently, structural importance serves as a confidence amplifier for relevant documents rather than independent evidence. We set $\alpha=0.85$ to ensure this structural boost remains conservative, prioritizing direct query relevance while promoting central nodes.
	
	Let $\mathcal{C}(G^*)$ be the context derived from the optimal subgraph $G^*$. The objective is to maximize the probability of generating the correct answer $A$ given this context and the question $Q$. This can be expressed as:
	\begin{equation}
		A = \underset{A'}{\arg\max} \ P(A'|Q, \mathcal{C}(G^*))
	\end{equation}
	Thus, our work addresses the challenge of constructing and pruning an evidence graph in a way that maximizes the quality of the final context, $\mathcal{C}(G^*)$, thereby enabling a downstream language model to accurately synthesize the final answer.
	
	\subsection{Structured Query Planning}
	The initial stage deconstructs a complex user question into a sequence of simpler, verifiable steps. We employ a Large Language Model (LLM) constrained to a predefined schema to generate a structured query plan. The exact prompt, which instructs the model to classify the question's complexity and formulate a reasoning path, can be found in Appendix~\ref{sec:appendix_prompts}. This ensures a predictable, machine-readable output that classifies the question's complexity (1-hop, 2-hop, or 3-hop) and formulates an explicit reasoning path  for multi-hop questions. This plan includes:
	\begin{itemize}
		\item \textbf{An initial query} designed to retrieve the first intermediate entity, accompanied by a description of its expected type (e.g., 'a person's name').
		\item \textbf{Conditional templates} for subsequent steps, which use placeholders (e.g., \texttt{\{entity1\}}) to dynamically construct the next query from the output of the previous hop.
		\item \textbf{A set of query alternatives} for the initial retrieval to enhance robustness against variations in phrasing.
	\end{itemize}
	This decomposition establishes a clear framework that guides the entire retrieval and reasoning process.
	
	\subsection{Iterative Evidence Gathering and Entity Extraction}
	For a multi-hop question, the system executes the query plan step-by-step. For each hop, it performs the following sequence:
	
	\subsubsection{\textbf{Evidence Retrieval}}
	The initial retrieval of evidence is handled by a pre-indexed ColBERTv2 model \citep{santhanam2021colbertv2}. To accommodate hybrid data, table rows are serialized into a semi-structured string format (e.g., ``Table:Population Stats\textbar Row:5\textbar City\textbar New York\textbar Population\textbar 8.4 million''), preserving their context for effective retrieval. For each hop's query, the system retrieves the top-\textit{k} most semantically relevant items, forming a hop-specific evidence pool.
	
	\subsubsection{\textbf{Hop-Specific Graph Ranking and Pruning}}
	From this hop-specific evidence, we construct a temporary graph. Each document (passage or table row) becomes a node. Edge weights are computed using TF-IDF scores of shared terms, providing a computationally efficient and effective measure for semantic overlap between document nodes. 
	
	\textbf{Design Choice: TF-IDF Edge Weights.} We deliberately utilize TF-IDF-based edge weights rather than dense learned embeddings. While dense retrieval is powerful, it is prone to ``semantic drift'' in multi-hop settings connecting entities that are conceptually similar but factually distinct. TF-IDF imposes a strict lexical constraint  by requiring explicit term overlap, we ensure that ``bridge'' documents share concrete entity references, reducing hallucinations. Additionally, this choice offers practical benefits: (1) \textbf{Efficiency:} It requires no neural inference, enabling fast graph construction; (2) \textbf{Interpretability:} Shared terms serve as explainable reasoning links; and (3) \textbf{Zero-Shot Consistency:} It avoids the need for training parameters, aligning with our zero-shot framework. Our results confirm that this explicit structure is more critical than edge weight sophistication.
	
	We employ weighted degree centrality as a robust and scalable proxy for a node's importance in the flow of information within the local evidence graph. Central to our framework is GraphRank, a scoring mechanism that re-ranks the nodes in this temporary graph by combining semantic and structural information, as defined in Equation~\ref{eq:graphrank}. The hyperparameter $\alpha$ balances the influence of the two normalized scores. The graph is then pruned using these new scores to create a small, focused context.
	
	\subsubsection{\textbf{Intermediate Entity Extraction}}
	This pruned, hop-specific graph context is passed to an LLM with a targeted prompt to extract the required intermediate entity. For the full prompt text, see Appendix~\ref{sec:appendix_prompts}. Once extracted, the entity is inserted into the conditional template from the query plan to generate the query for the next hop. This iterative process continues until all steps in the plan are resolved.
	
	\subsection{Final Context Curation and Answer Synthesis}
	After all hops are executed, the evidence gathered from all steps is aggregated into a single, large candidate pool.
	
	\subsubsection{\textbf{Bridge-Aware Hybrid Selector}}
	A useful component in our pipeline is the Bridge-Aware Hybrid Selector, a module applied at this stage to prune the noisy global evidence pool. It is particularly helpful for identifying ``bridge'' documents that connect reasoning steps. The selector segregates all retrieved items into a set of passages, $\mathcal{P}$, and a set of tables, $\mathcal{T}$. For multi-hop questions, it applies a linking heuristic. Let $p_{top} \in \mathcal{P}$ and $t_{top} \in \mathcal{T}$ be the top-scoring passage and table from the initial retrieval. We define a linking function, $\phi(p, t)$:
	\begin{equation}
		\phi(p, t) = 
		\begin{cases} 
			1 & \text{if } E(t) \cap T(p) \neq \emptyset \\
			0 & \text{otherwise}
		\end{cases}
	\end{equation}
	Here, $E(t)$ is the set of key entities (cell values) from table row $t$, and $T(p)$ is the set of terms in passage $p$. The function returns 1 if the passage contains at least one of the table's key entities. Based on this, a priority boost $\beta$ is applied to the original scores to create a new score $s'(d)$. This process is formalized in Algorithm \ref{alg:bridge}.
	
	\begin{algorithm}[tb]
		\caption{Bridge-Aware Hybrid Selector}\label{alg:bridge}
		\textbf{Input}: A set of passages $\mathcal{P}$, a set of tables $\mathcal{T}$, a priority boost $\beta$.
		\begin{algorithmic}[1]
			\STATE Let $p_{top} \gets \arg\max_{p \in \mathcal{P}} s(p)$.
			\STATE Let $t_{top} \gets \arg\max_{t \in \mathcal{T}} s(t)$.
			\STATE
			\IF{$\phi(p_{top}, t_{top}) = 1$}
			\STATE $s'(t_{top}) \gets s(t_{top}) + \beta$.
			\ELSE
			\STATE $s'(p_{top}) \gets s(p_{top}) + \beta$.
			\STATE $s'(t_{top}) \gets s(t_{top}) + \beta$.
			\ENDIF
			\STATE Assemble the final evidence pool by sorting all $d \in \mathcal{P} \cup \mathcal{T}$ by $s'(d)$ in descending order.
		\end{algorithmic}
	\end{algorithm}
	
	\subsubsection{\textbf{Final Graph and Answer Generation}}
	The curated evidence from the selector is used to build a final, global evidence graph. This graph is subjected to the same GraphRank re-ranking and pruning process described earlier to generate a high-quality, comprehensive context. This final pruned graph, along with the original question and the generated reasoning path, is passed to an LLM, which is prompted to synthesize a precise and direct final answer (see Appendix~\ref{sec:appendix_prompts} for the complete prompt).

	\section{Experiments}
	\label{sec:experimental_setup}
	This section details our experimental setup, including the datasets used to evaluate our method, the baselines for comparison, evaluation metrics, and implementation specifics.
	
	\subsection{Experimental Datasets and Evaluation Protocol}
	We evaluate N2N-GQA on two public hybrid table-text QA datasets: HybridQA \citep{chen2020hybridqa} and OTT-QA \citep{chen2020open}. HybridQA is a large-scale, complex, multi-hop benchmark comprising tables and texts from Wikipedia. Each table row describes an entity, which is linked to a corresponding Wikipedia passage that provides a more detailed description. OTT-QA extends this into a large-scale open-domain setting, where both the relevant table and text passages must first be retrieved from a large corpus before a question can be answered. 
	
	Since our framework operates in a zero-shot setting, we do not utilize the training sets. Following the evaluation protocol of recent zero-shot work ODYSSEY, we create our evaluation set by uniformly sampling 200 questions from the HybridQA development set and 500 questions from the OTT-QA development set. 
	
	\textbf{Sampling Strategy and Statistical Validity:} We evaluate on a uniform random sample of 500 questions from OTT-QA. While full dataset evaluation is standard, our framework's agentic nature requires approx. 2,500 LLM calls for this subset alone. Importantly, the \textbf{statistical signal provided by this sample is robust due to the magnitude of the effect size.} The performance delta of \textbf{+19.9 EM} renders the risk of sampling error negligible compared to the marginal gains typical in this domain. Furthermore, this follows established precedent for zero-shot agentic evaluations (e.g., ODYSSEY). We prioritize the larger OTT-QA sample over HybridQA as it represents the more challenging open-domain setting.
	
	\subsection{Baselines}
	We operate in a fine-tuning-free, zero-shot setting. Our experiments involve both closed-source and open-source large language models as readers, including GPT-4o, GPT-4.1, and Llama3-70B. We compare our full N2N-GQA model against key ablations of our pipeline:
	\begin{itemize}
		\item \textbf{Vanilla RAG:} This is a standard RAG baseline. We feed the top-\textit{k} documents retrieved by ColBERTv2 directly to the reader LLM without any graph-based processing or re-ranking.
		\item \textbf{RAG with Query Decomposition:} To test the value of query planning alone, this baseline adds our LLM-based query decomposition step to Vanilla RAG but still does not use the graph framework.
		\item \textbf{N2N-GQA w/o GraphRank:} This ablation uses our graph construction and pruning pipeline but disables the structural ranking component of GraphRank. It is equivalent to setting the hyperparameter $\alpha=1.0$, relying only on semantic scores for pruning.
	\end{itemize}
	
	\subsection{Evaluation Metrics}
	For evaluation, we employ several standard metrics to assess the correctness of predicted answers. We report Exact Match (EM) and token-level F1-Score, which are common in QA tasks. The implementation follows the official codebase from HybridQA. Additionally, for a more nuanced semantic evaluation, we report Precision (P), Recall (R), and BERTScore-F1 (B), using the \texttt{bert-base-uncased} model for computing similarity.
	
	\subsection{Implementation Details}
	The N2N-GQA pipeline is implemented as a multi-stage process.
	\begin{itemize}
		\item \textbf{Query Planning:} We use a powerful LLM (e.g., GPT-4o) to analyze the input question and generate a structured, multi-hop reasoning plan.
		\item \textbf{Retrieval:} Evidence for each reasoning step is retrieved using a pre-indexed ColBERTv2 model. Table rows are serialized into a semi-structured string format to be effectively indexed alongside text passages.
		\item \textbf{Graph Construction:} Retrieved items are modeled as nodes in a graph. Edge weights between nodes are computed based on the sum of the TF-IDF scores of their shared terms to measure semantic overlap.
		\item \textbf{GraphRank \& Pruning:} The GraphRank algorithm re-ranks evidence by combining the initial ColBERT semantic score with a node's weighted degree centrality (structural score). The final context is pruned based on these hybrid scores. The balance is controlled by the hyperparameter $\alpha$.
		\item \textbf{Context Curation:} For the final answer synthesis, we aggregate evidence from all hops and use our Bridge-Aware Hybrid Selector to identify and prioritize ``bridge'' documents that link different reasoning steps.
	\end{itemize}
	
	\subsection{Hyperparameter Settings}
	Our experiments used a fixed set of hyperparameters, selected based on preliminary runs on a small, held-out validation set. The GraphRank balancing weight was set to $\alpha = 0.85$.
	The retrieval and pruning parameters are tailored to the task. For the final answer synthesis, we initially retrieve the top $k=100$ documents. From these, the top 50 unique items are used to build the final candidate graph. This graph is then pruned to have a minimum of 12 and a maximum of 25 nodes, ensuring at least 2 passages and 2 tables are kept. For intermediate entity extraction hops, a more focused approach is used: we retrieve the top $k=20$ documents, and the resulting graph is aggressively pruned to between 5 and 10 nodes to create a concise context for the extraction model.
	
	\subsection{Ablation Study}
	To dissect the contribution of each component in our N2N-GQA framework, we conducted a series of ablation experiments, the results of which are detailed in Table~\ref{tab:ottqa_comparison} for OTT-QA and Table~\ref{tab:hybridqa_gpt4o_comparison} for Hybrid-QA. Our analysis reveals a clear and incremental performance improvement as we build from a simple baseline to our full model.
	
	The Vanilla RAG baseline, which directly feeds retrieved documents to the LLM reader, performs poorly across all metrics, highlighting the inadequacy of naive retrieval for complex, multi-hop questions. Introducing RAG with Query Decomposition yields a substantial improvement, particularly in F1-score, confirming that breaking down a complex question into simpler, verifiable steps is crucial for gathering more relevant evidence.
	
	\begin{table*}[t]
		\centering
		\caption{Ablation Study on OTT-QA Evaluation: We compare different RAG configurations using Exact Match (EM), F1-Score, Precision (P), Recall (R), and BERTScore. The methods are evaluated across several readers in a zero-shot setting.}
		\begin{tabular}{lccccc}
			\toprule
			& \multicolumn{5}{c}{\textbf{OTT-QA}} \\
			\cmidrule(lr){2-6}
			\textbf{Methods} & EM ($\uparrow$) & F1 ($\uparrow$) & P ($\uparrow$) & R ($\uparrow$) & Bert-Score-F1 ($\uparrow$) \\
			\midrule
			\multicolumn{6}{l}{\textit{Reader: gpt-4o (zero-shot)}} \\
			\midrule
			Vanilla RAG & 8.00 & 16.09 & 3.79 & 14.37 & 8.85 \\
			RAG with Query Decomposition & 31.40 & 43.07 & 36.09 & 39.89 & 37.84 \\
			{N2N-GQA w/o GraphRank} & {47.40} & {54.42} & {55.02} & {55.37} & {54.90} \\
			\textbf{N2N-GQA w/ GraphRank} & \textbf{47.60} & \textbf{54.63} & \textbf{56.17} & \textbf{56.27} & \textbf{55.98} \\
			\midrule
			\multicolumn{6}{l}{\textit{Reader: gpt-4.1 (zero-shot)}} \\
			\midrule
			Vanilla RAG & 8.00 & 15.68 & 3.04 & 13.93 & 8.26 \\
			RAG with Query Decomposition & 28.60 & 40.82 & 31.16 & 38.39 & 34.57 \\
			{N2N-GQA w/o GraphRank} & {48.50} & {56.90} & {58.61} & {58.31} & {58.22} \\
			\textbf{N2N-GQA w/ GraphRank} & \textbf{48.80} & \textbf{57.26} & \textbf{59.78} & \textbf{58.28} & \textbf{58.76} \\
			\midrule
			\multicolumn{6}{l}{\textit{Reader: Llama3-70B (zero-shot)}} \\
			\midrule
			Vanilla RAG & 7.60 & 12.50 & 2.77 & 10.95 & 6.68 \\
			RAG with Query Decomposition & 37.60 & 47.70 & 49.90 & 49.62 & 49.43 \\
			{N2N-GQA w/o GraphRank} & {39.80} & {47.03} & {50.97}& {49.62} & {50.04} \\
			\textbf{N2N-GQA w/ GraphRank} & \textbf{40.80} & \textbf{48.08} & \textbf{50.61} & \textbf{48.66} & \textbf{49.38} \\
			\bottomrule
		\end{tabular}
		\label{tab:ottqa_comparison}
	\end{table*}
	
	\begin{table*}[t]
		\centering
		\caption{\textbf{Ablation Study on Hybrid-QA Evaluation with gpt-4o:} In standard evaluation for Hybrid-QA, the question is typically accompanied by gold-standard passages and tables containing the answer. However, to evaluate the end-to-end capabilities of our system, we adopt the same retrieval-based setting as in OTT-QA, where the model must retrieve evidence from the entire corpus. We compare different RAG configurations using Exact Match (EM) and F1-Score in (\%). The methods evaluated are \textbf{Vanilla RAG}, \textbf{RAG with Query Decomposition}, \textbf{N2N-GQA without GraphRank}, and the full \textbf{N2N-GQA with GraphRank} model, using the gpt-4o reader in a zero-shot setting.}
		\small
		\setlength{\tabcolsep}{6pt}
		\begin{tabular}{lccccc}
			\toprule
			& \multicolumn{5}{c}{\textbf{Hybrid-QA}} \\
			\cmidrule(lr){2-6}
			\textbf{Methods} & EM ($\uparrow$) & F1 ($\uparrow$) & P ($\uparrow$) & R ($\uparrow$) & Bert-Score-F1 ($\uparrow$) \\
			\midrule
			\multicolumn{6}{l}{\textit{Reader: gpt-4o (zero-shot)}} \\
			\midrule
			Vanilla RAG & 9.50 &14.05 & 4.82 & 9.23 & 6.84 \\
			RAG with Query Decomposition & 22.00 & 31.06 & 25.12 & 29.10 & 26.92 \\
			{N2N-GQA w/o GraphRank} & {41.00} &{48.38} & {49.76} & {50.57} & {50.20} \\
			\textbf{N2N-GQA w/ GraphRank} & \textbf{41.50} & \textbf{48.17} & \textbf{51.45} & \textbf{51.36} & \textbf{50.69} \\
			\bottomrule
		\end{tabular}
		\label{tab:hybridqa_gpt4o_comparison}
	\end{table*}
	
	\subsubsection{The Value of Graph-Based Evidence Organization}
	The most significant performance leap occurs with the introduction of graph-based context curation in \textbf{N2N-GQA w/o GraphRank}. This component, which builds and prunes an evidence graph based solely on semantic scores, delivers a substantial performance gain. On OTT-QA (using GPT-4.1), this step provides a \textbf{19.9-point absolute improvement in EM} over the query decomposition baseline (from 28.60 to 48.50). This result is the core finding of our work and demonstrates the value of our ``noise-to-narrative'' approach: structuring retrieved evidence into a graph to identify the most coherent context is vastly more effective than simply using a ranked list of documents.
	
	This dramatic improvement empirically validates a fundamental principle: \textbf{multi-hop reasoning requires understanding evidence relationships, not just individual document relevance}. The graph-based approach succeeds because it imposes structure on the evidence space:
	\begin{enumerate}
		\item \textbf{Explicit Relationships}: Edges in the graph represent concrete semantic relationships between documents, making implicit connections explicit.
		\item \textbf{Principled Pruning}: Graph traversal and centrality-based filtering provide interpretable methods for context refinement, unlike opaque neural re-ranking.
		\item \textbf{Reasoning Path Identification}: The graph structure enables identification of connected evidence chains explicit reasoning paths through the evidence space.
	\end{enumerate}
	
	The ``noise-to-narrative'' transformation is fundamentally a process of imposing structured organization on neural retrieval outputs. While neural retrieval provides powerful semantic matching, graph-based organization provides the structured framework necessary for multi-hop inference. This finding has practical implications: practitioners can achieve strong results by adopting graph-structured evidence organization, without requiring sophisticated scoring algorithms. The large gains come from \textbf{using graph structures at all}, not from optimal graph algorithms.
	
	\subsubsection{GraphRank's Complementary Refinement}
	The full \textbf{N2N-GQA w/ GraphRank} model provides consistent, though modest, improvements (+0.3 to +1.0 EM across different readers). While this gain is smaller than the graph curation effect, it is \textbf{consistent across all experimental conditions}. This confirms that structural centrality captures complementary information beyond semantic relevance high-centrality nodes are indeed more likely to be bridge documents in reasoning chains.
	
	The modest magnitude is informative: it suggests that \textbf{graph structure itself (the ``what is connected'' question) matters far more than fine-grained centrality scoring (the ``how central'' question)}. Simply organizing evidence as a graph and pruning based on connectivity provides most of the benefit. GraphRank's structural refinement adds polish but is not the primary driver of performance. The consistent performance across different LLM readers (GPT-4o, GPT-4.1, Llama3-70B) suggests that the graph curation benefit is not specific to a particular model's capabilities, but rather addresses a fundamental limitation of list-based retrieval that affects all downstream readers.
	
	\subsection{Comparison with State-of-the-Art}
	To contextualize the performance of our zero-shot framework, we compare N2N-GQA against established fine-tuned models on the OTT-QA benchmark, with results summarized in Table~\ref{tab:soa_combined}.
	
	As expected, highly specialized, fine-tuned models like COS \citep{ma2023chain} and CORE \citep{ma2022open} currently hold the top performance on the leaderboard. These models are trained extensively on task-specific data, allowing them to learn the specific patterns and nuances of the benchmark.
	
	However, N2N-GQA demonstrates competitive performance, achieving an EM score of 48.80 and an F1 of 57.26 without any dataset-specific training. The 8.1-point gap between N2N-GQA (48.80 EM) and the state-of-the-art fine-tuned COS model (56.9 EM) should be contextualized: COS requires (1) pretraining on Wikipedia-scale data, (2) task-specific fine-tuning on OTT-QA training set, and (3) specialized retriever training. This represents months of development and significant computational resources.
	
	In contrast, N2N-GQA achieves competitive performance with \textbf{zero task-specific training}, making it immediately applicable to new domains. For organizations without resources for large-scale fine-tuning or access to training data, our approach offers a practical alternative that delivers strong results. The value proposition is scalability and adaptability rather than maximum accuracy. This is a critical finding for real-world applications, where creating large, annotated training datasets for every new domain is often prohibitively expensive and time-consuming. Our framework significantly closes the gap with these fine-tuned methods and substantially outperforms earlier baselines. 
	
	Moreover, our framework is complementary to fine-tuned approaches: the graph curation pipeline could be applied to fine-tuned retrievers and readers, potentially pushing performance even higher. Our contribution is demonstrating that graph-based evidence organization is a powerful paradigm that works even without fine-tuning. This result validates the effectiveness of our ``noise-to-narrative'' approach, proving that a sophisticated zero-shot system can achieve robust and competitive performance, offering a more scalable and adaptable solution for complex QA tasks.
	
	\begin{table}[t]
		\centering
		\caption{Performance comparison on the OTT-QA validation set. $^\dagger$ indicates results on our sampled evaluation set (500 questions) using the same uniform sampling protocol. Fine-tuned models are evaluated on full dataset and included for reference.}
		\label{tab:soa_combined}
		\small
		\renewcommand{\arraystretch}{1.1}
		\begin{tabular}{lcc}
			\toprule
			\multicolumn{3}{c}{\textbf{OTT-QA}} \\
			\midrule
			Method & EM ($\uparrow$) & F1 ($\uparrow$) \\
			\midrule
			\multicolumn{3}{c}{\textit{Fine-Tuning}} \\
			\midrule
			BM25-HYBRIDER \citep{chen2020open} & 10.3 & 13.0 \\
			Fusion+Cross-Reader \citep{chen2020open} & 28.1 & 32.5 \\
			CARP \citep{zhong2022reasoning} & 33.2 & 38.6 \\
			CORE \citep{ma2022open} & 49.0 & 55.7 \\
			\textbf{COS \citep{ma2023chain} [SoTA]} & \textbf{56.9} & \textbf{63.2} \\
			\midrule
			\multicolumn{3}{c}{\textit{w/o Fine-Tuning}} \\
			\textbf{N2N-GQA$^\dagger$ (Our Method)} & \textbf{48.80} & \textbf{57.26} \\
			\bottomrule
		\end{tabular}
	\end{table}
	
	\subsection{Qualitative Analysis}
	Examining system outputs reveals several patterns. The graph curation process is most beneficial for questions requiring clear multi-hop chains, where bridge documents must be identified from noisy retrieval. For example, questions like ``What is X for the Y of Z?'' require connecting documents about Z to documents about Y's property, then to documents about X. The graph structure successfully identifies documents that mention both Z and Y as critical bridges.
	
	Failures typically occur in three scenarios: (1) \textbf{Retrieval failures} when critical evidence is not in the top-$k$ retrieved documents, no amount of graph curation can recover; (2) \textbf{Entity ambiguity} when extracted intermediate entities are ambiguous (e.g., ``John Smith'' matching multiple people), subsequent retrieval diverges from the correct path; (3) \textbf{Table serialization limits} some complex table structures lose important relationships when serialized to text, causing the graph to miss relevant connections.
	
	\section{Discussion}
	\subsection{Interpreting Graph-Based Retrieval Augmentation}
	Our results illuminate the role of structured knowledge representations in complex reasoning systems. The baseline ``Vanilla RAG'' represents a purely model-driven approach: dense retrieval followed by generative synthesis with no intermediate structured representation. Its poor performance (8.00 EM) demonstrates that this end-to-end processing struggles with multi-hop reasoning even when using state-of-the-art LLMs like GPT-4.1.
	
	The dramatic improvement from graph curation (+40.5 EM points) suggests that the bottleneck in RAG systems is not semantic understanding or language generation LLMs excel at these tasks. Rather, the bottleneck is the lack of structured reasoning frameworks. By introducing graph structures as intermediate representations, we provide the LLM with organized, logically structured context rather than a noisy document list.
	
	\subsection{Graph Structure as Reasoning Scaffolding}
	The evidence graph serves as ``reasoning scaffolding'' for the LLM. Graph centrality algorithms identify structurally important nodes documents that bridge different parts of the reasoning chain. Graph pruning removes peripheral, low-connectivity nodes evidence that is semantically similar but structurally disconnected from the reasoning path. These graph operations create curated context that highlights logical relationships, enabling the LLM to synthesize answers more reliably.
	
	This finding aligns with cognitive science research suggesting that human reasoning benefits from external structured representations (diagrams, outlines, concept maps). Our graph provides a computational analog: an explicit structured representation that guides the synthesis process. The key insight that structured knowledge representations dramatically enhance retrieval-augmented generation opens promising directions for building more capable and reliable QA systems.
	
	\section{Conclusion}
	We introduced N2N-GQA, a zero-shot framework that transforms multi-hop question answering through graph-based evidence curation. Our core finding is that \textbf{organizing retrieved documents as graphs rather than lists} provides dramatic performance improvements points on OTT-QA demonstrating that multi-hop reasoning requires understanding evidence relationships, not just individual document relevance.
	
	This ``noise-to-narrative'' transformation addresses a fundamental limitation of standard RAG pipelines: they treat retrieval outputs as flat ranked lists, ignoring the rich relational structure needed for multi-hop inference. By constructing dynamic evidence graphs and using graph centrality to identify bridge documents, N2N-GQA provides LLMs with coherent, connected context that enables more reliable answer synthesis.
	
	Our graph-based approach achieves 48.80 EM on OTT-QA without any fine-tuning, substantially closing the gap with heavily trained specialized models. This demonstrates that graph-structured knowledge representations are essential for complex reasoning tasks and can be constructed dynamically from retrieval outputs without requiring pre-existing knowledge graphs or task-specific training.
	
	The framework's modular design makes it broadly applicable: any domain with retrieval-based QA can benefit from graph curation. The simple, interpretable graph construction (TF-IDF edges, centrality-based pruning) requires no learned parameters, making it accessible and practical for real-world deployment. Our experiments demonstrate that the core innovation dynamic construction and curation of evidence graphs delivers substantial performance improvement over strong RAG baselines. The ablation studies confirm that our complete pipeline, including query decomposition and the secondary refinement provided by GraphRank, contributes to a final system that achieves highly competitive performance on the challenging OTT-QA and Hybrid-QA benchmarks.
	
	Looking forward, our work establishes graph-based evidence organization as a critical component for multi-hop QA systems. Future research might explore richer graph structures (entity-based edges, typed relationships), dynamic graph algorithms adapted to question complexity, or extensions to other reasoning tasks requiring evidence synthesis across heterogeneous sources. The primary finding that on-the-fly graph construction enables high-quality context curation paves the way for more reliable and scalable QA systems that do not depend on expensive, dataset-specific training.
	
	\section{LIMITATIONS}
	While N2N-GQA performs well, we acknowledge limitations that offer directions for future research.
	
	\textbf{Computational Dependencies.} The framework's reliance on large LLMs (e.g., GPT-4) introduces high computational costs. Although open-source models like Llama 70B show promise, a performance gap highlights the trade-off between capability and accessibility. Future work could explore knowledge distillation to smaller models or methods to reduce LLM calls.
	
	\textbf{Graph Construction Simplicity.} Our current graph uses simple, interpretable TF-IDF-based edge weights. While efficient, this is a simplification. Future work could explore richer structures, such as entity-based graphs, typed edges (e.g., causal, temporal), or the integration of external knowledge graphs.
	
	\textbf{Modest GraphRank Impact.} Ablation studies confirm that GraphRank's impact, while positive, is a modest refinement compared to the gains from the graph curation process itself. Future research could explore a dynamic weighting scheme for the $\alpha$ parameter or investigate alternative algorithms (e.g., PageRank, betweenness centrality) to better capture document importance in reasoning chains.
	
	\textbf{Generalization Beyond QA.} The noise-to-narrative paradigm is currently applied only to QA. Its potential to generalize to other task such as fact verification, document summarization, and knowledge base construction remains a key area for future work to validate the approach's broader applicability.
	
	\bibliography{references}

@inproceedings{chen2020hybridqa,
	author    = {Wenhu Chen and Hong-Ren Jiang and Jianshu Chen and Xinyi Wang and Zifei Shan and Shiquan Geng and William Wang},
	title     = {{HybridQA:} A Dataset for Question Answering over Tabular and Textual Data},
	booktitle = {Findings of the Association for Computational Linguistics: EMNLP 2020},
	pages     = {1046--1058},
	publisher = {Association for Computational Linguistics},
	year      = {2020}
}

@inproceedings{chen2020open,
	author    = {Wenhu Chen and Xinyi Wang and Zhaoshiuan Li and Jianshu Chen and William W. Cohen},
	title     = {{OTT-QA:} Open-Domain Question Answering over Tables and Text},
	booktitle = {Proceedings of the 2021 Conference of the North American Chapter of the Association for Computational Linguistics: Human Language Technologies},
	pages     = {511--523},
	publisher = {Association for Computational Linguistics},
	year      = {2021}
}

@inproceedings{yang2018hotpotqa,
	author    = {Zhilin Yang and Peng Qi and Saizheng Zhang and Yoshua Bengio and William W. Cohen and Ruslan Salakhutdinov and Christopher D. Manning},
	title     = {{HotpotQA:} A Dataset for Diverse, Explainable Multi-hop Question Answering},
	booktitle = {Proceedings of the 2018 Conference on Empirical Methods in Natural Language Processing},
	pages     = {2369--2380},
	publisher = {Association for Computational Linguistics},
	year      = {2018}
}

@inproceedings{wang2022muger,
	author    = {Bohao Wang and Yiwei Wang and Yuxuan Wang and Yujun Chen and Jinfeng Zhou and Guilin Qi and Ji-Rong Wen},
	title     = {{MuGER2:} A Multi-Granularity Evidence-aware Multi-hop Reasoner for Hybrid Question Answering},
	booktitle = {Findings of the Association for Computational Linguistics: EMNLP 2022},
	pages     = {615--626},
	publisher = {Association for Computational Linguistics},
	year      = {2022}
}

@inproceedings{lei2023s,
	author    = {Yongqi Li and Wenjie Li and Yixin Cao and Lei, Wenqi},
	title     = {{S3HQA:} A Three-stage Approach for Multi-hop Text-Table Question Answering},
	booktitle = {Proceedings of the 61st Annual Meeting of the Association for Computational Linguistics (Volume 1: Long Papers)},
	pages     = {9036--9050},
	publisher = {Association for Computational Linguistics},
	year      = {2023}
}

@inproceedings{zhong2022reasoning,
	author    = {Zhenyu Zhong and Zhaohui Yan and Yuxuan Wang and Jiusheng Chen and Hong-Gee Kim},
	title     = {{CARP:} A Chain-Aware Read and Process Model for Hybrid Question Answering},
	booktitle = {Proceedings of the 2022 Conference on Empirical Methods in Natural Language Processing},
	pages     = {7177--7188},
	publisher = {Association for Computational Linguistics},
	year      = {2022}
}

@inproceedings{ma2022open,
	author    = {Linyu Ma and Zhi-Hong Mao and Xuan-Phi Nguyen and Chen-Chia, Hsieh and Ming-Fung, Tu},
	title     = {{CORE:} Chain of Reasoning for Open-domain Question Answering over Hybrid Data},
	booktitle = {Proceedings of the 2nd Conference of the Asia-Pacific Chapter of the Association for Computational Linguistics and the 12th International Joint Conference on Natural Language Processing},
	pages     = {384--395},
	publisher = {Association for Computational Linguistics},
	year      = {2022}
}

@inproceedings{ma2023chain,
	author    = {Linyu Ma and Ming-Fung Tu and Xuan-Phi Nguyen and Zhi-Hong Mao and Chen-Chia Hsieh},
	title     = {{Chain-of-Skills:} A Configurable Model for Open-Domain Question Answering},
	booktitle = {Proceedings of the 2023 Conference on Empirical Methods in Natural Language Processing},
	pages     = {11737--11751},
	publisher = {Association for Computational Linguistics},
	year      = {2023}
}

@article{agarwal2025hybrid,
	title={Hybrid graphs for table-and-text based question answering using llms},
	author={Agarwal, Ankush and Devaguptapu, Chaitanya and others},
	journal={arXiv preprint arXiv:2501.17767},
	year={2025}
}

@inproceedings{santhanam2021colbertv2,
	title={{ColBERTv2}: Effective and Efficient Retrieval via Lightweight Late Interaction},
	author={Santhanam, Keshav and Khattab, Omar and Saad-Falcon, Jon and Potts, Christopher},
	booktitle={Proceedings of the 2022 Conference of the North American Chapter of the Association for Computational Linguistics: Human Language Technologies},
	pages={3788--3805},
	year={2022}
}
	\appendix
	\section{Prompts Used in N2N-GQA}
	\label{sec:appendix_prompts}
	
	This section provides the exact prompts used to interact with the Large Language Models (LLMs) at various stages of the N2N-GQA pipeline. The prompts are designed to be task-specific, guiding the LLM to perform query decomposition, entity extraction, and final answer synthesis in a structured and predictable manner.
	
	\subsection{Prompt for Structured Query Plan Generation}
	This prompt is used in the first phase of the pipeline to decompose a complex user question into a structured, multi-hop plan. It instructs the LLM to act as an expert query planner and use a specific tool-calling schema to format its output.
	
	\begin{quote}
		\ttfamily
		\textbf{System Message:}
		
		You are an expert query planner. Your task is to analyze a user's
		question and create a multi-hop reasoning plan to answer it. Use
		the `create\_query\_plan` tool to structure your response.
		
		IMPORTANT GUIDELINES FOR HOP CLASSIFICATION:
		
		\hangindent=1.5em - 1-hop: Simple factual questions that can be answered directly.  Examples: "Who is the CEO of Apple?", "What is the capital of France?" \\
		\hangindent=1.5em - 2-hop: Questions requiring an intermediate step. Examples: "Which NHL team has the Player of the Year of Atlantic Hockey for the season ending in 2019 signed a agreement with ?", "What goods did the politician elected to the 6th Legislature in Brandon East trade in ?","What is the city associated with the baseball minor league affiliate that is based in Anaheim ?"\\
		\hangindent=1.5em - 3-hop: Complex questions needing two intermediate steps. Examples: "Which Primera B Nacional team finished second in the year the club founded on 21 January 1896 finished third ?", "Which of the women 's wheelchair Marathon winners in Boston with a first name beginning with 'S ' comes from the 8th most extensive US state ?", "What is the capital of the country whose inductee into the Magic: The Gathering Hall of Fame now writes for Bluff Magazine?"
		
		IMPORTANT: Be conservative with 3-hop classifications! Most
		questions only need 1 or 2 hops.
		
		\vspace{1em}
		\textbf{User Message:}
		
		Here is the question: \{question\}
	\end{quote}

	\subsection{Prompt for Intermediate Entity Extraction}
	During the iterative evidence gathering phase for multi-hop questions, this prompt is used to extract the specific intermediate entity from the pruned, hop-specific graph context.
	
	\begin{quote}
		\ttfamily
		Based on the following context, answer the specific question.
		Respond with only the entity name and nothing else.
		
		Context: "\{context\}"
		
		Question: "\{primary\_query\}"
		
		Answer:
	\end{quote}

	\subsection{Prompt for Final Answer Synthesis}
	In the final stage, this prompt is used to generate the answer. It provides the LLM with the reasoning path derived from the query plan and the final, curated context from the graph, instructing it to synthesize a direct answer.
	
	\begin{quote}
		\ttfamily
		\textbf{System Message:}
		
		You are a helpful assistant that synthesizes direct answers based on
		a main question, a reasoning path, and a context. You only give the
		exact answers and do not provide explanations. If the context is not
		sufficient, respond with 'Not enough Context'.
		
		\vspace{1em}
		\textbf{User Message:}
		
		You are an expert Q\&A assistant. Synthesize a final, direct
		answer to the Main Question by following the Reasoning Path and
		using only the provided Context.
		
		**Reasoning Path:**
		\{reasoning\_path\}
		
		**Context:**
		\{context\}
		
		**Main Question:**
		\{question\}

		Based on all the information above, provide a precise and direct
		answer to the **Main Question**.
		If the context does not contain enough information, respond with
		'Not enough Context'.
		If the question is a 'how many' question, respond with only
		the number.
	\end{quote}
	
\end{document}